%% file: paper.tex
\documentclass[11pt]{article}
\usepackage{coling2020}
\usepackage{times}
\usepackage{url}
\usepackage{latexsym}

\usepackage{amsmath,amssymb,amsfonts}
\usepackage{algorithmic}
\usepackage{graphicx}
\usepackage{textcomp}
\usepackage{xcolor}

\usepackage{booktabs} 
\usepackage{wrapfig}
\usepackage{bbm}
\usepackage{xspace}
\usepackage{pifont}

\usepackage{hyperref}
\usepackage{xcolor}
\usepackage{paralist}
\usepackage{subcaption}
\usepackage{enumitem}
\usepackage{multirow}
\usepackage{footmisc}

\usepackage{tabularx}
    \newcolumntype{L}{>{\raggedright\arraybackslash}X}
\input{dfn}

\hypersetup{
	colorlinks,
	citecolor={black},
	urlcolor={blue}
}

\colingfinalcopy 

\title{Auto-Encoding Variational Bayes for Inferring Topics and Visualization}

\author{Dang Pham, Tuan M. V. Le \\
  Department of Computer Science \\
  New Mexico State University \\
  {\tt \{dangpnh, tuanle\}@nmsu.edu} }

\date{}

\begin{document}
	
\maketitle
\input{abstract}

\blfootnote{
	%
	%
	\hspace{-0.65cm}  
	This work is licensed under a Creative Commons 
	Attribution 4.0 International License.
	License details:
	\url{http://creativecommons.org/licenses/by/4.0/}.
	%
	%
	%
	%
}
\input{intro}
\input{related}
\input{aevb}
\input{experiments}
\input{conclusion}

\section*{Acknowledgements}
This research is sponsored by NSF \#1757207 and NSF \#1914635.

\newpage
\bibliographystyle{coling}
\bibliography{bib}

\end{document}

%% file: dfn.tex
\newcommand{\lda}{{\em LDA}}
\newcommand{\plda}{{\em ProdLDA-VAE}}
\newcommand{\tsne}{{\em t-SNE}}
\newcommand{\largevis}{{\em LargeVis}}
\newcommand{\plsv}{{\em PLSV}}

\newcommand{\ldavae}{{\em LDA-VAE}}

\newcommand{\cbit}{\begin{compactitem}}
	\newcommand{\ceit}{\end{compactitem}}
\newcommand{\cben}{\begin{compactenum}}
	\newcommand{\ceen}{\end{compactenum}}

\definecolor{OliveGreen}{rgb}{0,0.6,0}

\newcommand{\bit}{\begin{itemize}}
	\newcommand{\eit}{\end{itemize}}
\newcommand{\ben}{\begin{enumerate}}
	\newcommand{\een}{\end{enumerate}}
\newcommand{\beq}{\begin{equation}}
\newcommand{\eeq}{\end{equation}}

\newcommand{\mD}{\mathcal{D}}
\newcommand{\mV}{\mathcal{V}}
\newcommand{\mN}{\mathcal{N}}

\newcommand{\mR}{\mathcal{R}}
\newcommand{\mL}{\mathcal{L}}
\newcommand{\mE}{\mathbb{E}}
\newcommand{\mhD}{\mathbb{D}}
\newcommand{\mhR}{\mathbb{R}}
\newcommand{\mhZ}{\mathbb{Z}}

\newcommand{\news}{\textsc{20 Newsgroups}}
\newcommand{\reuters}{\textsc{Reuters}}
\newcommand{\wos}{\textsc{Web of Science}}
\newcommand{\arxiv}{\textsc{Arxiv}}

\newcommand{\plsvmap}{{\em PLSV-MAP}}
\newcommand{\plsvvae}{{\em PLSV-VAE}}

\newcommand{\bbeta}{\boldsymbol{\beta}}
\newcommand{\bTheta}{\boldsymbol{\Theta}}
\newcommand{\bX}{\boldsymbol{X}}

\newcommand{\bPhi}{\boldsymbol{\Phi}}
\newcommand{\bUpsilon}{\boldsymbol{\Upsilon}}

\newcommand{\hide}[1]{}

%% file: abstract.tex
\begin{abstract}

Visualization and topic modeling are widely used approaches for text analysis. Traditional visualization methods find low-dimensional representations of documents in the visualization space (typically 2D or 3D) that can be displayed using a scatterplot. In contrast, topic modeling  aims to discover topics from text, but for visualization, one needs to perform a post-hoc embedding using dimensionality reduction methods. Recent approaches propose using a generative model to jointly find topics and visualization, allowing the semantics to be infused in the visualization space for a meaningful interpretation. A major challenge that prevents these methods from being used practically is the scalability of their inference algorithms. We present, to the best of our knowledge, the first fast Auto-Encoding Variational Bayes based inference method for jointly inferring topics and visualization. Since our method is black box, it can handle model changes efficiently with little mathematical rederivation effort. We demonstrate the efficiency and effectiveness of our method on real-world large datasets and compare it with existing baselines.

\end{abstract}

%% file: intro.tex
\section{Introduction}
Visualization and topic modeling are important tools in the analysis of text corpora. Visualization methods, such as \tsne\ \cite{maaten2008visualizing}, find low-dimensional representations of documents in the visualization space (typically 2D or 3D) that can be displayed using a scatterplot. Such visualization is useful for exploratory tasks. However, there is a lack of semantic interpretation as those visualization methods do not extract topics. In contrast, topic modeling  aims to discover semantic topics from text, but for visualization, one needs to perform a post-hoc embedding using dimensionality reduction methods. Since this pipeline approach may not be ideal, there has been recent interest in jointly inferring topics and visualization using a single generative model \cite{iwata2008probabilistic}. This joint approach allows the semantics to be infused in the visualization space where users can view documents and their topics. The problem of jointly inferring topics and visualization can be formally stated as follows.




\noindent \textbf{Problem.} Let $\mD=\{\mathbf{w}_n\}_{n=1}^{\mN}$ denote a finite set of $\mN$ documents and let $\mV$ be a finite vocabulary from these documents.
Given a number of topics $Z$, and visualization dimension $d$, we want to find:
\begin{compactitem}
	\item For topic modeling: $Z$ latent topics, and their word distributions collectively denoted as $\bbeta=\{\beta_z\}_{z=1}^{Z}$, topic distributions of documents collectively denoted as $\bTheta={\{\theta_n\}_{n=1}^{\mN}}$, and
	\item For visualization: $d$-dimensional visualization coordinates for $\mN$ documents $\bX = \{x_n\}_{n=1}^{\mN}$, and $Z$ topics $\bPhi=\{\phi_z\}_{z=1}^{Z}$ such that the distances between documents, topics in the visualization space reflect the topic-document distributions $\bTheta$.
\end{compactitem}

To solve this problem, \plsv\ (Probabilistic Latent Semantic Visualization) is the first model that attempts to tie together all latent variables of topics and visualization (i.e.,  $\bUpsilon=\{\bX, \bPhi, \bbeta\}$) in a generative model. Its tight integration between visualization and the underlying topic model can support applications such as user-driven topic modeling where users can interactively provide feedback to the model \cite{choo2013utopian}. \plsv\ can also be used as a basic building block when developing new models for other analysis tasks, such as visual comparison of document collections \cite{le2019contravis}.

Relatively less attention has been paid to methods for fast inference of topics and visualization. Existing models often use Maximum a Posteriori (MAP) estimation with the EM algorithm, which is difficult to scale to large datasets. As shown in Figure \ref{fig:run_time}, to run a \plsv\ model of 50 topics via MAP estimation on a dataset of modest size (e.g., \news), it takes more than 18 hours using a single core. This long running time limits the usability of these visualization methods in practice.

In this paper, we aim to propose a fast Auto-Encoding Variational Bayes (AEVB) based inference method for inferring topics and visualization. AEVB \cite{aevb} is a black-box variational method which is efficient for inference and learning in latent Gaussian Models with large datasets. However, to apply the AEVB approach to topic models like \lda, one needs to deal with problems caused by the Dirichlet prior and by posterior collapse \cite{he19iclr}. One of the successful AEVB based methods proposed to tackle those challenges for topic models is AVITM \cite{Srivastava2017AutoencodingVI}.


It is not straightforward to apply AEVB or AVITM to our problem because of two main challenges. First, as reviewed in Section \ref{related}, \plsv\ models a document's  topic distribution using a softmax function over its Euclidean distances to topics. It is not clear how to express this nonlinear functional relationship between three categories of latent variables (i.e., topic distribution $\theta_n$, document coordinate $x_n$, and topic coordinate $\phi_z$) when applying AVITM to visualization. Second, AEVB has an assumption that latent encodings are identically and independently distributed (i.i.d.) across samples \cite{casale2018gaussian} \cite{LinCBTR19}. In our case, this assumption works well with latent document coordinates $\bX$ where each document $n$ is associated with its latent encoding $x_n$ in the visualization space. However, for topic coordinates $\bPhi$ and word probabilities $\bbeta$, that assumption is too strong. The reason is that latent encodings of any topic $k$ w.r.t any documents are not independent, but in fact, in our extreme case these latent encodings are similar, i.e., $\phi_z^{(i)}=\phi_z^{(j)}$, for any documents $i, j$ and any topic $z$. In other words, $\phi_z$ is shared across documents. The same argument also applies to word probabilities $\bbeta$.


To address the first challenge, we propose to model the nonlinear functional relationship between $\theta_n$, $x_n$, $\bPhi$ using a normalized Radial Basis Function (RBF) Neural Network \cite{bishop1995neural}. In this model, $\phi_z \in \bPhi$ is the center vector for neuron $z$, i.e., $\bPhi$ are treated as parameters of the RBF network and will be estimated. Similarly, we model $\bbeta$ as parameters of a linear neural network that is connected to the RBF network to form the decoder in the AEVB approach. By treating $\bPhi$ and $\bbeta$ as parameters of the decoder, we can solve the second challenge, though it can be seen that our algorithm does not learn their posterior distributions but rather their point estimates. In Section \ref{sec:model}, we present in detail our proposed method. We focus on \plsv\ model in this work, though the proposed AEVB inference method could be easily adapted to other visualization models. 

We summarize our contributions as follows:
\begin{compactitem}
    \item We propose, to the best of our knowledge, the first AEVB inference method for the problem of jointly inferring topics and visualization.
    \item In our approach, we design a decoder that includes an RBF network connected to a linear neural network. These networks are parameterized by topic coordinates and word probabilities, ensuring that they are shared across all documents. 
    \item We conduct extensive experiments on real-world large datasets, showing the efficiency and effectiveness of our method. While running much faster than \plsv, it gains better visualization quality and comparable topic coherence.
    
    \item Since our method is black box, it can handle model changes efficiently with little mathematical rederivation effort. We implement different \plsv\ models that use different RBFs by just changing a few lines of code. We experimentally show that \plsv\ with Gaussian or Inverse quadratic RBFs consistently produces good performance across datasets.
\end{compactitem}

%% file: related.tex
\section{Background and Related Work} \label{related}
\subsection{Topic Modeling and Visualization}
Topic models \cite{blei2003latent,Hofmann1999ProbabilisticLS} are widely used for unsupervised representation learning of text and have found applications in different text mining tasks \cite{ramage2009topic,blei2007correlated,tkachenko2019comparelda,kim2019topicsifter}. Popular topic models such as \lda\ \cite{blei2003latent}, find a low-dimensional representation of each document in topic space. Each dimension of the topic space has a meaning attached to it and is modeled as a probability distribution over words. In contrast,
 \tsne\ \cite{maaten2008visualizing}, \largevis\ \cite{tang2016visualizing} are visualization methods aiming to find for each document a low-dimensional representation (typically 2D or 3D). However, we often do not have such semantic interpretation for that low-dimensional space as in topic models. Therefore, there have been works attempting to infuse semantics to the visualization space by jointly modeling topics and visualization \cite{iwata2008probabilistic,le2014manifold}. These methods often suffer from the scalability issue with large datasets. In this work, we aim to scale up these methods by proposing a fast AEVB based inference method. We focus on \plsv\  \cite{iwata2008probabilistic} for applying our proposed method. \plsv\ has been used as a basic block for building new models for visual text mining tasks \cite{le2014semantic,le2019contravis}. Our proposed method could be easily adapted to these models. 

\plsv\ assumes the following process to generate documents and visualization:
\begin{compactenum}
	\item For each topic $z=1,\cdots, Z$:
	\begin{compactenum}
		\item Draw a word distribution: $\beta_{z} \sim \operatorname{Dirichlet}(\lambda)$
		\item Draw a topic coordinate: $\phi_{z} \sim \operatorname{Normal}\left(\mathbf{0}, \varphi \boldsymbol{I}\right)$
	\end{compactenum}
	\item  For each document $n=1, \cdots, \mN$:
	\begin{compactenum}
		\item Draw a document coordinate: $x_{n} \sim \operatorname{Normal}\left(\mathbf{0}, \gamma \boldsymbol{I}\right)$
		\item For each word $w_{nm}$ in document $n$:    
		\begin{compactenum}
			\item Draw a topic: $z \sim \operatorname{Multi}\left(\left\{p\left(z | x_{n}, \mathbf{\Phi}\right)\right\}_{z=1}^{Z}\right)$
			\item Draw a word: $w_{nm} \sim \operatorname{Multi}\left( \beta_z \right)$ 
		\end{compactenum}
	\end{compactenum}
\end{compactenum}


Here $\beta_z$ has a Dirichlet prior. Topic and document coordinates have Gaussian priors of the forms: $p(\phi_z|\varphi)= \left(\frac{1}{2\pi \varphi}\right)^{d/2} \exp (-\frac{\left\|\phi_{z}\right\|^{2}}{2\varphi})$ and $p(x_n|\gamma)= \left(\frac{1}{2\pi \gamma}\right)^{d/2} \exp (-\frac{\left\|x_{n}\right\|^{2}}{2\gamma})$ respectively. The topic distribution of a document is defined using a softmax function over its distances to topics:
\vspace{-0.1in}
{
\begin{equation} \label{eq:softmax}
\theta_{nz} = p\left(z | x_{n}, \mathbf{\Phi}\right)=\frac{\exp \left(-\frac{1}{2}\left\|x_{n}-\phi_{z}\right\|^{2}\right)}{\sum_{z^{\prime}=1}^{Z} \exp \left(-\frac{1}{2}\left\|x_{n}-\phi_{z^{\prime}}\right\|^{2}\right)}
\vspace{-0.1in}
\end{equation}
}
As we can see from Eq. \ref{eq:softmax}, the $z$th topic proportion of document $n$ is high when document coordinate $x_n$ is close to topic coordinate $\phi_z$. This relationship ensures that the distances between documents, topics in the visualization space reflect the topic-document distributions $\bTheta$. In the \plsv\ paper, the parameters $\bUpsilon=\{\bX, \bPhi, \bbeta\}$ are estimated using MAP estimation with the EM algorithm. As shown in our experiments, the algorithm does not scale to large datasets.

\subsection{Auto-Encoding Variational Bayes for Topic Models}
AEVB \cite{Kingma2014AutoEncodingVB} and its variant WiSE-ALE \cite{LinCBTR19}, AVITM \cite{Srivastava2017AutoencodingVI} are black-box variational inference methods whose purpose is to allow practitioners to quickly explore and adjust the model's assumptions with little rederivation effort \cite{Ranganath2014BlackBV}. AVITM is an auto-encoding variational inference method for topic models. It approximates the true posterior $p(\theta, \boldsymbol{z}|\mathbf{w},\alpha,\beta)$ using a variational distribution $q(\theta, \boldsymbol{z}|\mathbf{w}, \eta,\rho)$ where $\alpha$ is hyperparameter of Dirichlet prior and $\eta,\rho$ are the free variational parameters over $\theta,\boldsymbol{z}$ respectively. Different from Mean-Field Variational Inference, AVITM computes the variational parameters using an inference neural network and they are chosen by optimizing the following ELBO (i.e., the lower bound to the marginal log likelihood):
\begin{equation}
\mL\left(\eta,\rho | \alpha, \beta \right)= -\mhD_{\mathrm{KL}}\left[q(\theta, \boldsymbol{z}|\mathbf{w}, \eta,\rho) \| p(\theta, \boldsymbol{z}|\alpha)\right] + \mE_{q(\theta, \boldsymbol{z}|\mathbf{w}, \eta,\rho)} \left[ \log p(\mathbf{w}|\theta,\boldsymbol{z},\alpha,\beta) \right]
\end{equation}

By collapsing $\boldsymbol{z}$ and approximating the Dirichlet prior $p(\theta|\alpha)$ with a logistic normal distribution, the second term (i.e., the expectations with respect to $q$) in the ELBO can be approximated using the reparameterization trick as in AEVB. The second term is also referred to as an expected negative reconstruction error in variational auto-encoders (VAE). While AVITM is successfully applied to \lda, it is not straightforward to apply it to our problem as discussed in the introduction.

%% file: aevb.tex
\section{Proposed Auto-Encoding Variational Bayes for Inferring Topics and Visualization} \label{sec:model}
We represent a document $n$ as a row vector of word counts:  $\mathbf{w}_n \in \mhZ_{\geq}^{|\mV|}$ and $\mathbf{w}_{n}^{v}$ is the number of occurrences of word $v \in \mV$ in the document. The marginal likelihood of a document is given by:
{\small
\begin{equation} \label{likelihood}
\hspace{-0.05in}
p\left(\mathbf{w}_n|\gamma, \bPhi, \bbeta\right) = \int_{x}\left( \prod_{v=1}^{|\mV|} \left(\sum_{z=1}^{Z} p(v|z,\bbeta)p\left(z|x, \mathbf{\Phi}\right)\right)^{\mathbf{w}_{n}^{v}}\right)p(x|\gamma)dx = \int_{x}\left( \prod_{v=1}^{|\mV|} p\left(v|x, \mathbf{\Phi},\bbeta\right)^{\mathbf{w}_{n}^{v}}\right)p(x|\gamma)dx
\end{equation}
}
The marginal likelihood of the corpus is $p(\mD) = \prod_{n=1}^{\mN}p\left(\mathbf{w}_n|\gamma, \bPhi, \bbeta\right)$. Note that here we treat $\bPhi$, and $\bbeta$ as fixed quantities that are to be estimated. Therefore we are working with a non-smoothed \plsv\ where $\bPhi$, and $\bbeta$ are not endowed with a posterior distribution. By treating $\bPhi$, and $\bbeta$ as model parameters, we ensure that they are shared across all documents in the AEVB approach. We will consider a fuller Bayesian approach to \plsv\ in our future work.

As in AVITM, we collapse the discrete latent variable $\boldsymbol{z}$ to avoid the difficulty of determining a reparameterization function for it. The rightmost integral in Eq. \ref{likelihood} is the marginal likelihood after $\boldsymbol{z}$ is collapsed. We now only consider the true posterior distribution over latent variable $x$: $p\left(x|\mathbf{w}_n,\gamma, \bPhi, \bbeta \right)$. Due to the intractability of Eq. \ref{likelihood}, it is intractable to compute the posterior. We approximate it by a variational distribution $q\left(x|\mathbf{w}_n, \eta\right)$ parameterized by $\eta$. The variational parameter $\eta$ is estimated using an inference network as in AEVB. We have the following lower bound to the marginal log likelihood (ELBO) of a document:
\begin{equation}\label{plsv_elbo}
\mL\left(\eta| \gamma, \bPhi, \bbeta \right)= -\mhD_{\mathrm{KL}}\left[q(x|\mathbf{w}_n,\eta) \| p(x|\gamma)\right] + \mE_{q(x|\mathbf{w}_n,\eta)} \left[ \log p\left(\mathbf{w}_n|x, \bPhi, \bbeta \right) \right]
\end{equation}

Since the prior $p(x|\gamma) = \operatorname{Normal}\left(\mathbf{0}, \gamma \boldsymbol{I}\right)$ is a Gaussian, we can let the variational posterior $q(x|\mathbf{w}_n, \eta)$ be a Gaussian with a diagonal covariance matrix: $q(x|\mathbf{w}_n, \eta) = \operatorname{Normal}\left(\boldsymbol{\mu}_n, \boldsymbol{\Sigma}_n\right)$. The KL divergence between two Gaussians in Eq. \ref{plsv_elbo} can be computed in a closed form as follows \cite{kalai2010efficiently}:
\begin{equation}\label{eqn:kl}
\mhD_{\mathrm{KL}}\left[q(x|\mathbf{w}_n,\eta) \| p(x|\gamma)\right] =
\frac{1}{2}\left(\operatorname{tr}\left((\gamma\boldsymbol{I})^{-1} \boldsymbol{\Sigma}_{n}\right)+\left(-\boldsymbol{\mu}_{n}\right)^{\top} (\gamma\boldsymbol{I})^{-1}\left(-\boldsymbol{\mu}_{n}\right)-d+\log \frac{|\gamma\boldsymbol{I}|}{|\boldsymbol{\Sigma}_{n}|}\right)
\end{equation}

where $\boldsymbol{\mu}_n$, diagonal $\boldsymbol{\Sigma}_n \in \mhR^d$ are outputs of the encoding feed forward neural network with variational parameters $\eta$. The expectation w.r.t $q(x|\mathbf{w}_n,\eta)$ in Eq. \ref{plsv_elbo} can be estimated using reparameterization trick \cite{aevb}. More specifically, we sample $x^{(l)}$ from the posterior $q(x|\mathbf{w}_n,\eta)$ by using reparameterization over random variable $x$, i.e., $x^{(l)}=\boldsymbol{\mu}_n+\boldsymbol{\Sigma}_n^{1/2}\boldsymbol{\epsilon}^{(l)}$ where $\boldsymbol{\epsilon}^{(l)}\sim \operatorname{Normal}\left(\mathbf{0}, \boldsymbol{I}\right)$. The expectation can then be approximated as:
\vspace{-0.15in}
\begin{equation}\label{eq:recon}
\mE_{q(x|\mathbf{w}_n,\eta)} \left[ \log p\left(\mathbf{w}_n|x, \bPhi, \bbeta \right) \right] \approx \frac{1}{L}\sum_{l=1}^{L} \log p\left(\mathbf{w}_n|x^{(l)}, \bPhi, \bbeta \right)
\vspace{-0.15in}
\end{equation}

In Eq. \ref{eq:recon}, the decoding term $\log p\left(\mathbf{w}_n|x^{(l)}, \bPhi, \bbeta \right)$ is computed as:
\begin{equation}\label{eq:decodeterm}
\log p\left(\mathbf{w}_n|x^{(l)}, \bPhi, \bbeta \right) = \log \left(\theta_n^{(l)} \boldsymbol{\beta} \right)\mathbf{w}_n^T
\end{equation}

where $\boldsymbol{\beta} \in \mhR^{Z \times V}$ is the topic-word probability matrix, $\mathbf{w}_n \in \mR^{|\mV|}$ is a row vector of word counts, $\theta_n^{(l)} \in \mhR^Z$ is a row vector of topic proportions and $\theta_{nz}^{(l)}=p\left(z | x^{(l)}, \mathbf{\Phi}\right)$ is computed as in Eq. \ref{eq:softmax}. 
Based on Eq. \ref{eq:decodeterm} and Eq. \ref{eq:softmax}, we propose using a decoder with two connected neural networks:

\noindent \textbf{Normalized Radial Basis Function Network for computing $\theta_{nz}$.} We generalize $\theta_{nz}$ in Eq. \ref{eq:softmax} using a Normalized Radial Basis Function (RBF) Network \cite{bishop1995neural} as follows:
\begin{equation}\label{eqn:rbf}
	\theta_{nz}^{(l)}=p\left(z | x^{(l)}, \mathbf{\Phi}\right) = \frac{\sum_{z'=1}^{Z}w_{z,z'}\rho(\left\|x-\phi_{z'}\right\|)}{\sum_{z'=1}^{Z}\rho(\left\|x-\phi_{z'}\right\|)}
\end{equation}

In this network, we have $Z$ neurons in the hidden layer and $\phi_{z'}$ is the center vector for neuron $z'$. The RBF function $\rho$ is a non-linear function that depends on the distance $\left\|x-\phi_{z'}\right\|$ and $w_{z,z'}$ is the influence weight of neuron $z'$ on $\theta_{nz}$ where $\sum_{z'=1}^{Z}w_{z,z'}=1$. While $w_{z,z'}$ can be estimated by optimizing the ELBO, we choose to fix it as $w_{z,z'}=1$ when $z=z'$ and 0 otherwise. The parameters of this network are then the center vectors of $Z$ neurons that are the coordinates of topics in the visualization space. The RBF function $\rho$ can have different forms, e.g., Gaussian: $\exp(-\frac{1}{2}r^2)$, Inverse quadratic: $\frac{1}{1+r^2}$, or Inverse multiquadric: $\frac{1}{\sqrt{1+r^2}}$ where $r=\left\|x-\phi_{z'}\right\|$\footnote{{$r$ is Euclidean distance in our experiments}}. When $\rho$ is Gaussian, Eq. \ref{eqn:rbf} reduces to Eq. \ref{eq:softmax}. Note that this generalization of $\theta_{nz}$ is also discussed in \cite{le2016semantic} but not in the context of VAE inference. Since topic coordinates $\phi_{z'}$ are now the parameters of the RBF network, they can be shared and used by all documents for computing the topic distributions $\theta_n^{(l)}$. In the experiments, we will show the performance of \plsv\ with these RBFs using VAE inference. 

\begin{figure*}
	\centering
    \includegraphics[width=1.0\textwidth]{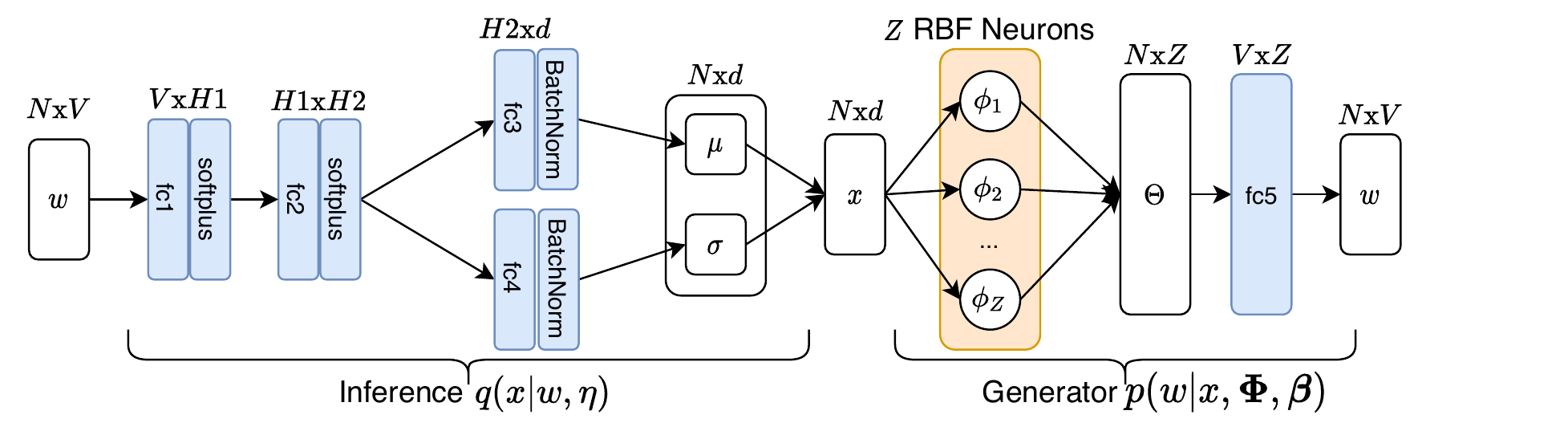}
    \vspace{-0.15in}
    \caption{The architecture of Variational Auto-Encoder for Visualization and Topic Modeling.}
    \label{fig:network}
    \vspace{-0.2in}
\end{figure*}

\noindent \textbf{Linear Neural Network for computing $\left(\theta_n^{(l)} \boldsymbol{\beta} \right)$.} The output of the above normalized RBF network will be the input of a linear neural network to compute $\left(\theta_n^{(l)} \boldsymbol{\beta} \right)$ in the decoding term. We treat $\boldsymbol{\beta}$ as the parameters, i.e., the linear weights $W$, of the network and it is computed using a softmax over the network weights to ensure the simplex constraint on $\boldsymbol{\beta}$: $\boldsymbol{\beta}=\sigma(W)$. The architecture of the whole Variational Auto-Encoder is given in Figure \ref{fig:network}. We use batch normalization \cite{pmlr-v37-ioffe15} to mitigate the posterior collapse issue found in the AEVB approach \cite{he19iclr,RazaviOPV19}.

\vspace{0.1in}
\noindent \textbf{Final Variational Objective Function.} From Eqs. \ref{plsv_elbo}, \ref{eqn:kl}, \ref{eq:recon}, \ref{eq:decodeterm}, we have the following objective function:
{\small
\begin{equation}
\begin{split}
\mL\left(\Omega\right) = \sum_{n=1}^{\mN} \Bigg[-\frac{1}{2}\left(\operatorname{tr}\left((\gamma \boldsymbol{I})^{-1} \boldsymbol{\Sigma}_{n} \right) + \left(-\boldsymbol{\mu}_{n} \right)^{T} (\gamma \boldsymbol{I})^{-1} \left(-\boldsymbol{\mu}_{n} \right) - d + \log \frac{|\gamma\boldsymbol{I}|}{|\boldsymbol{\Sigma}_{n}|} \right) + \frac{1}{L}\sum_{l=1}^{L} \log \left(\theta_n^{(l)} \boldsymbol{\beta} \right)\mathbf{w}_n^T \Bigg]
\end{split}
\end{equation}
}
where $\Omega=\{\bX, \bPhi, \bbeta,\eta\}$ represents all model and variational parameters, $\theta_{nz}^{(l)}=p\left(z | x^{(l)}\right)$ (Eq. \ref{eqn:rbf}), $\boldsymbol{\beta}=\sigma(W)$, $x^{(l)}=\boldsymbol{\mu}_n+\boldsymbol{\Sigma}_n^{1/2}\boldsymbol{\epsilon}^{(l)}$ and $\boldsymbol{\epsilon}^{(l)}\sim \operatorname{Normal}\left(\mathbf{0}, \boldsymbol{I}\right)$.


%% file: experiments.tex
\section{Experiments}
\input{exp1}
\subsection{Topic Coherence}
We quantitatively measure the quality of topic models produced by all methods in terms of topic coherence. The objective is to show that while having better visualization quality, \plsvvae\ also gains comparable, if not better, topic coherence. For topic coherence evaluation, we use Normalized Pointwise Mutual Information (NPMI) which has been shown to be correlated with human judgments \cite{lau2014machine}. NPMI is computed as follows:
\vspace{-0.1in}
\begin{equation}
NPMI(w_i,w_j)= \frac{\log \frac{p(w_i,w_j)}{p(w_i)p(w_j)}}{-\log p(w_i,w_j)}
\end{equation}

We estimate $p(w_i,w_j)$, $p(w_i)$, and $p(w_j)$ using Wikipedia 7-gram dataset\footnote{{https://nlp.cs.nyu.edu/wikipedia-data/}} created from the Wikipedia dump data as of June 2008 version. NPMI of a topic is computed as an average of the pairwise NPMI of its top 10 words. For each method, we average NPMI of its topics. Figure \ref{fig:npmi} shows topic coherence NPMI of all methods. As we can see, \plsvvae\ finds topics as good as those found by other methods, and in some settings, \plsvvae\ can find significantly better topics. For a qualitative evaluation of topic quality, we show some example topics found by \plsvvae\ in Figure \ref{fig:arxiv_Z50}.

\subsection{Visualization Examples} \label{exp:visualization}
\begin{figure}
	\centering
	\begin{subfigure}[b]{0.4\textwidth}
	\includegraphics[width=\textwidth]{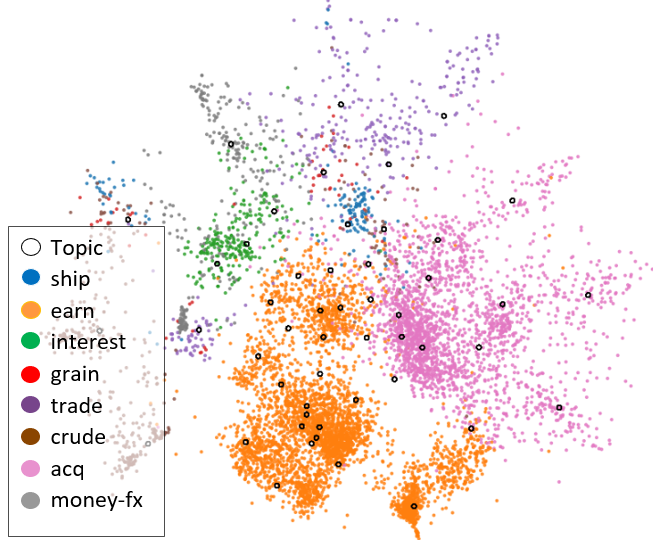}
	\subcaption{\reuters\ (\plsvmap)}
	\end{subfigure}
	\begin{subfigure}[b]{0.4\textwidth}
	\includegraphics[width=\textwidth]{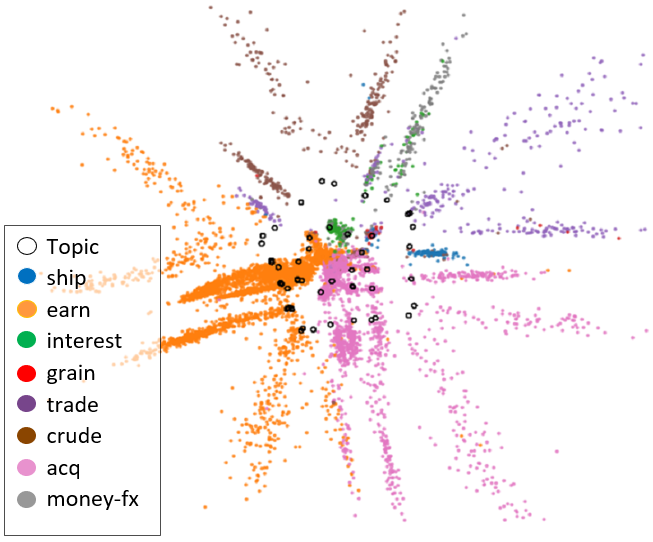}
	\subcaption{\reuters\ (\plsvvae)}
	\end{subfigure}
	
	\caption{Visualization of \reuters\ by a) \plsvmap\ b) \plsvvae.}
	\label{fig:reuters_comparison}
	\centering
	\begin{subfigure}[b]{0.4\textwidth}
	\includegraphics[width=\textwidth]{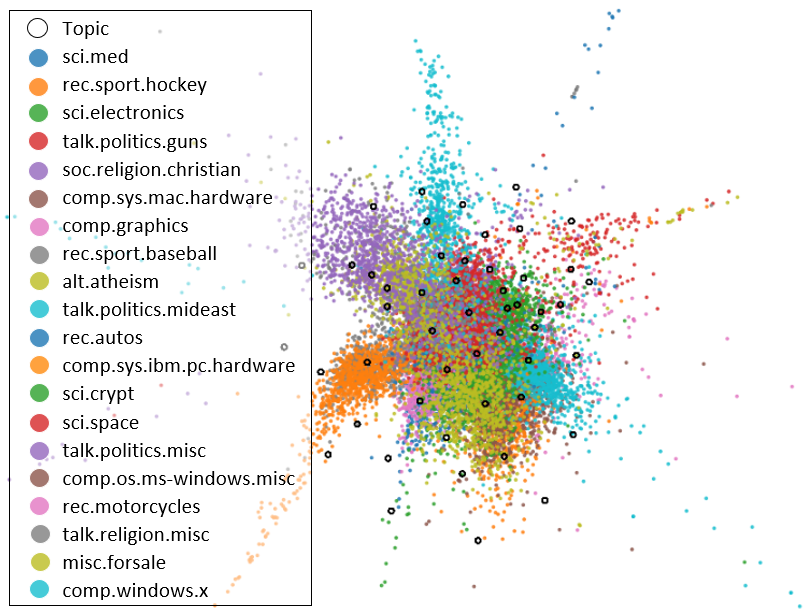}
	\subcaption{\news\ (\plsvmap)}
	\end{subfigure}
	\begin{subfigure}[b]{0.4\textwidth}
	\includegraphics[width=\textwidth]{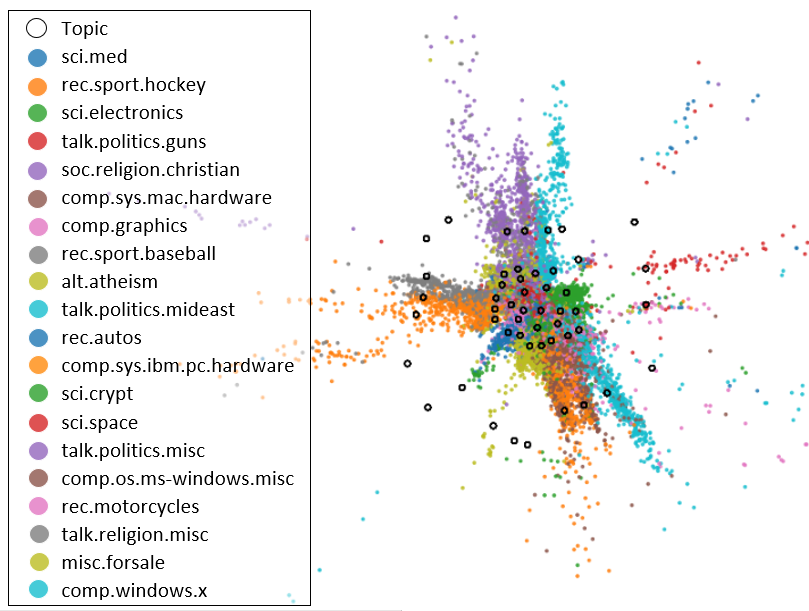}
	\subcaption{\news\ (\plsvvae)}
	\end{subfigure}
	\caption{Visualization of \news\ by a) \plsvmap\ b) \plsvvae.}
	\label{fig:20news_comparison}
	\centering
	\begin{subfigure}[b]{0.4\textwidth}
	\includegraphics[width=\textwidth]{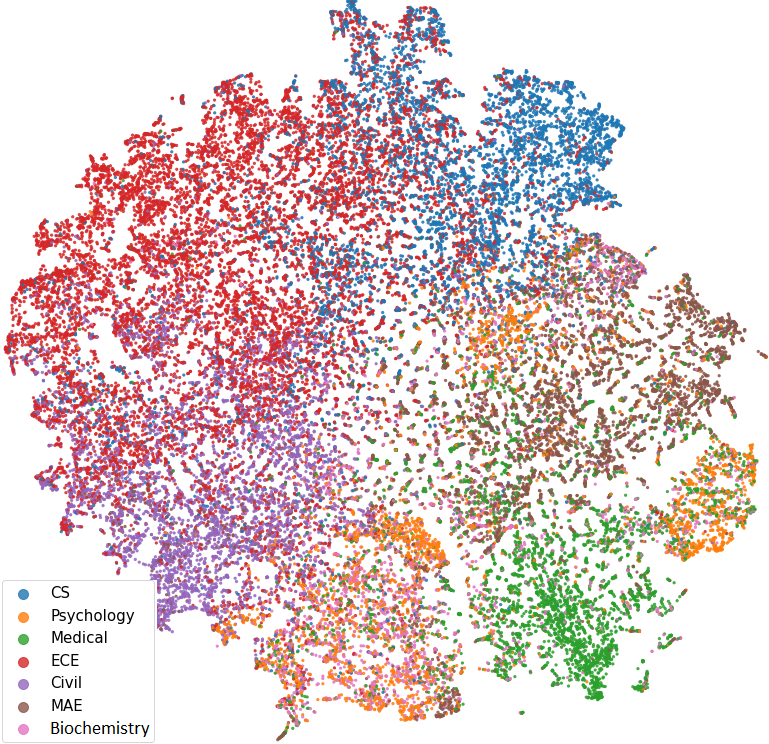}
	\subcaption{\wos\ (\plda\ + \tsne)}
	\end{subfigure}
	\begin{subfigure}[b]{0.4\textwidth}
	\includegraphics[width=\textwidth]{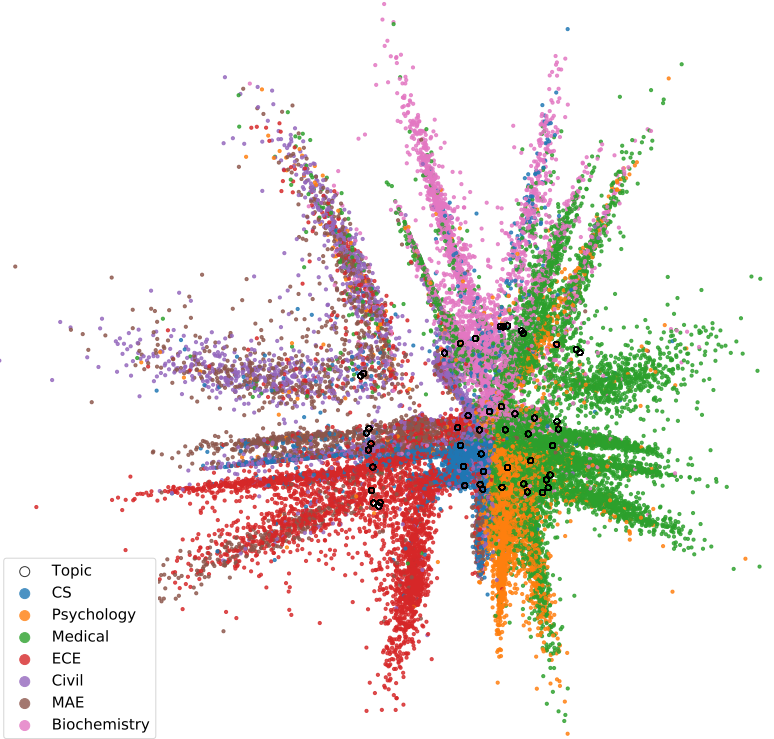}
	\subcaption{\wos\ (\plsvvae)}
	\end{subfigure}
	\caption{Visualization of \wos\ by a) \plda\ + \tsne\ b) \plsvvae.}
	\label{fig:mendeley_comparison}
\end{figure}

\begin{figure}
	\centering
	\begin{subfigure}[b]{0.4\textwidth}
	\includegraphics[width=\textwidth]{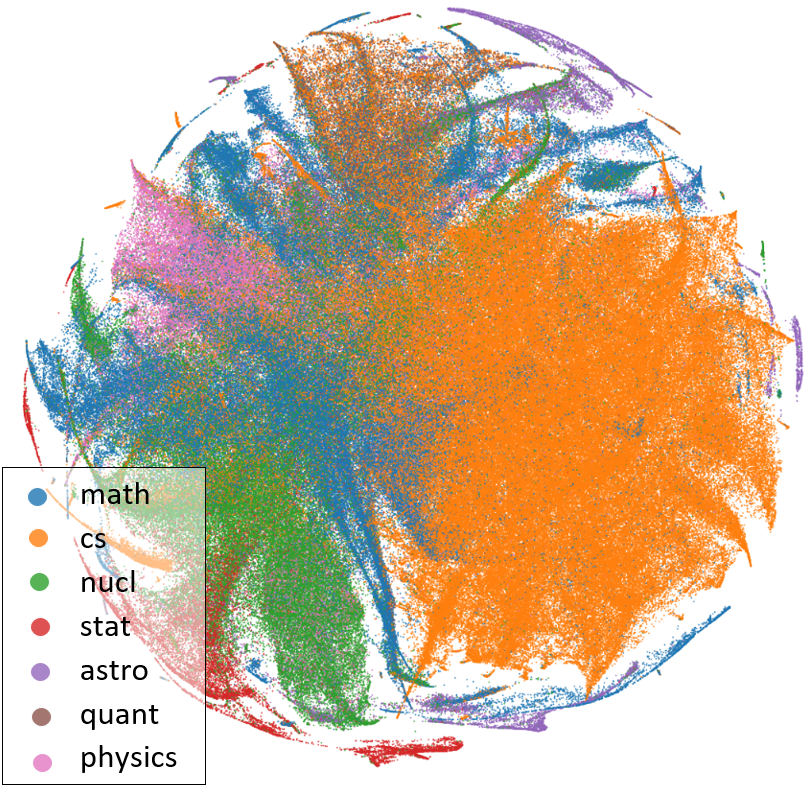}
	\subcaption{\arxiv\ (\plda\ + \tsne)}
	\end{subfigure}
	\begin{subfigure}[b]{0.4\textwidth}
	\includegraphics[width=\textwidth]{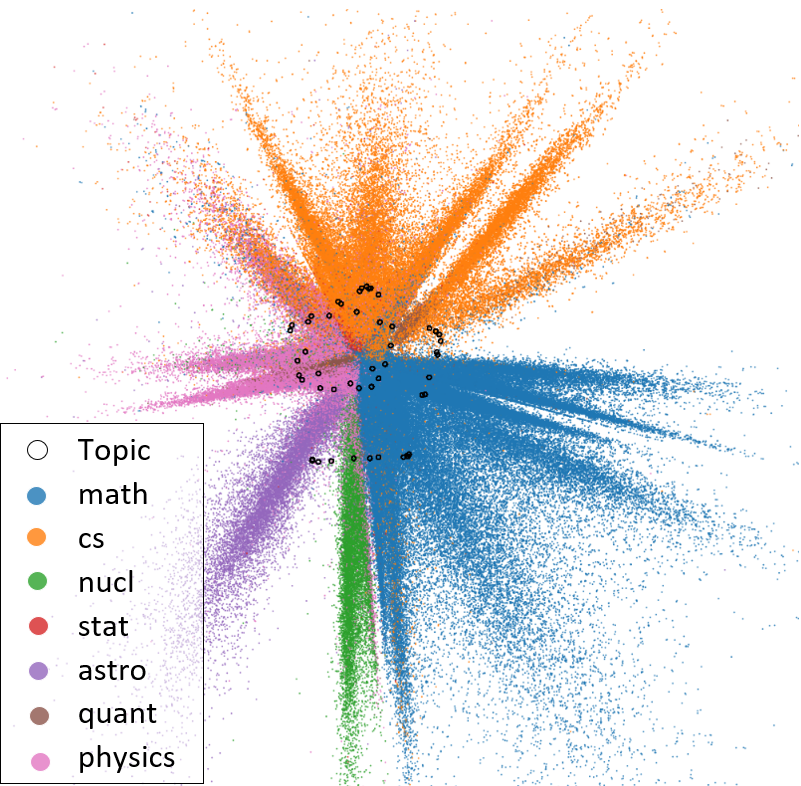}
	\subcaption{\arxiv\ (\plsvvae)}
	\end{subfigure}
	\caption{Visualization of \arxiv\ by a) \plda\ + \tsne\ b) \plsvvae.}
	\label{fig:Arxiv_comparison}
\vspace*{\floatsep}
	\centering
	\includegraphics[width=1.0\textwidth]{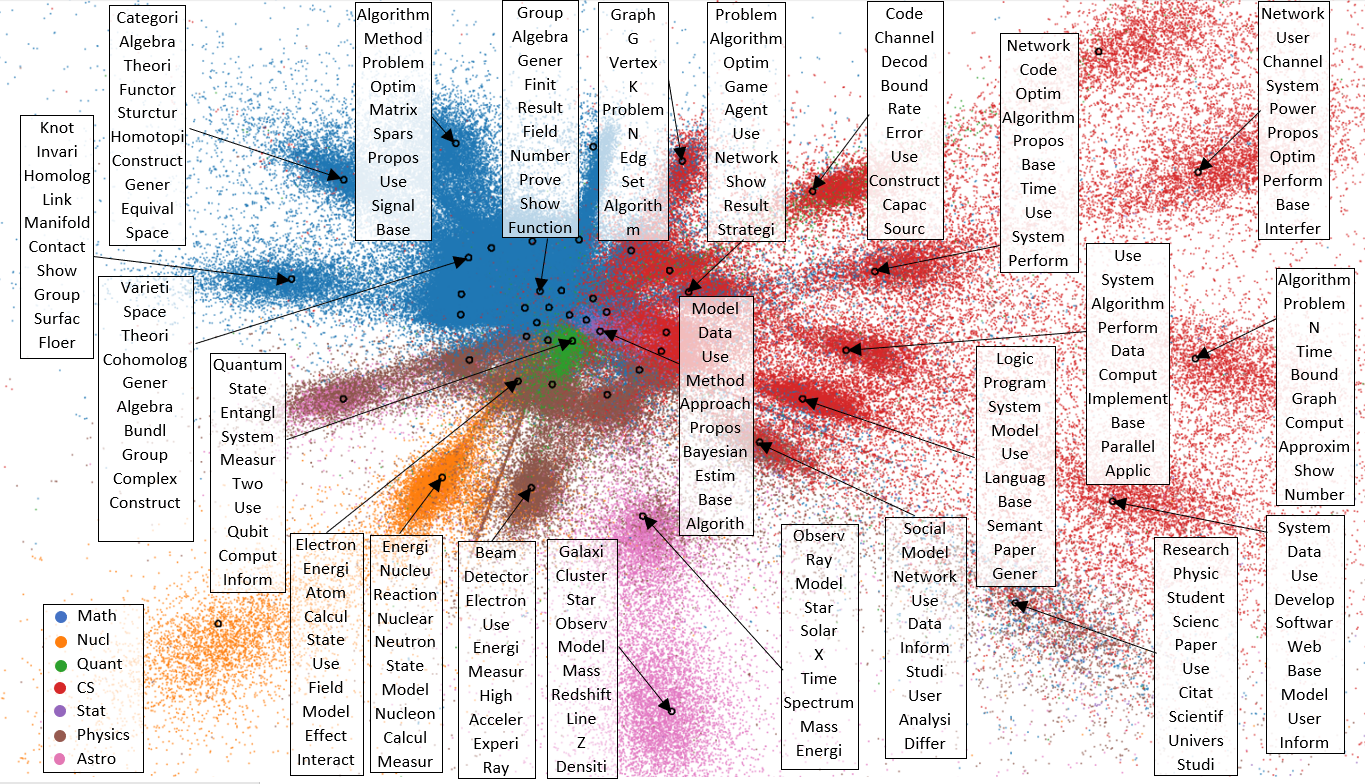}
	\caption{Visualization and topics found by \plsvvae\ (Inverse quadratic) on \arxiv\ ($Z = 50$ topics).}
	\label{fig:arxiv_Z50}

\vspace*{\floatsep}
\includegraphics[width=\textwidth]{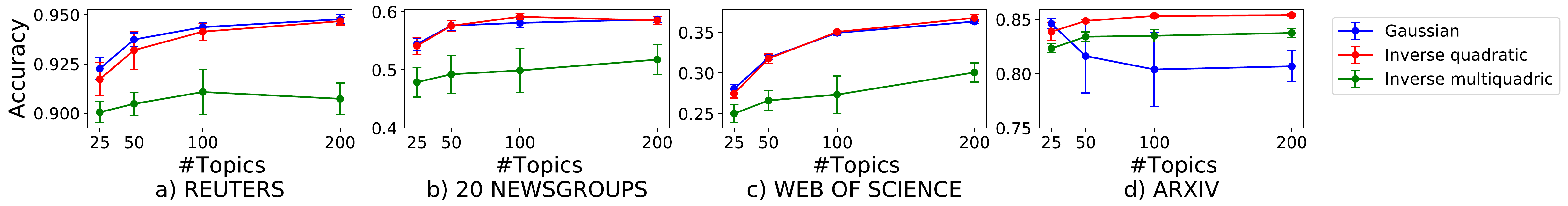}
\caption{$k$-NN accuracy in the visualization space by \plsvvae\ with different RBFs ($k=10$).}
\label{fig:KNNRBF_KNN10}

\vspace*{\floatsep}

\includegraphics[width=\textwidth]{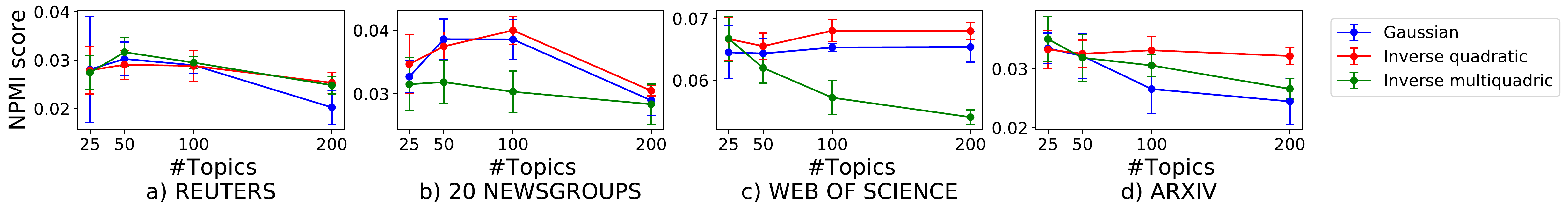}
\caption{Topic coherence NPMI by \plsvvae\ with different RBFs (vary number of topics $Z$).}
\label{fig:npmi_rbf}

\end{figure}

\begin{figure}

\includegraphics[width=\textwidth]{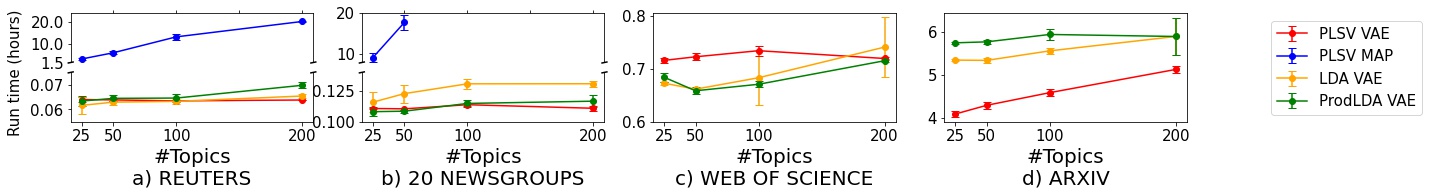}
\caption{Running time comparison.}
\label{fig:run_time}

\end{figure}
We compare visualizations produced by all methods qualitatively by showing some visualization examples.
In these visualizations, each document is represented by a point and the color of each point indicates the class of that document.
Figures \ref{fig:reuters_comparison} and \ref{fig:20news_comparison} present visualizations by \plsvmap, \plsvvae\ on \reuters\ and \news.
We see that \plsvvae\ can find meaningful clusters of documents.
For example, \plsvvae\ in Figure \ref{fig:reuters_comparison}(b) separates well the eight classes into different clusters such as the pink cluster for \textit{acq}, the orange cluster for \textit{earn}, and the brown cluster for \textit{crude}.
The visualization by \plsvmap\ in Figure \ref{fig:reuters_comparison}(a) also shows clear clusters but it runs much slower than \plsvvae\ as shown in Section \ref{exp:topics}. Figure \ref{fig:20news_comparison} presents visualization outputs for \news. For this more challenging dataset, \plsvvae\ produces better-separated clusters, as compared to \plsvmap. For example, \textit{baseball} and \textit{hockey} are mixed in Figure \ref{fig:20news_comparison}(a) by \plsvmap\ but these classes are separated better in Figure \ref{fig:20news_comparison}(b) by \plsvvae.
We do not show visualizations of \wos\ and \arxiv\ by \plsvmap\ because it fails to return any results even after 24 hours of running. We instead show visualizations of these two large datasets by \plsvvae\ and \plda\ + \tsne\ in Figures \ref{fig:mendeley_comparison} and \ref{fig:Arxiv_comparison}. As we can see, visualizations by \plsvvae\ are more intuitive than the ones by \plda\ + \tsne, which supports the outperformance of the joint approach over the pipeline approach.
\subsection{Comparing Different Radial Basis Functions} \label{exp:rbf}
\hide{
\begin{figure*}
	\includegraphics[width=\textwidth]{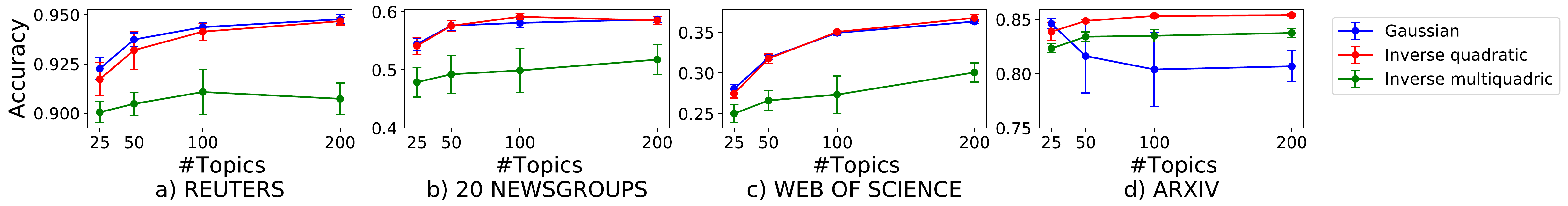}
	\caption{$k$-NN accuracy in the visualization space by \plsvvae\ with different RBFs ($k=10$).}
	\label{fig:KNNRBF_KNN10}
\end{figure*}

\begin{figure*}
	\includegraphics[width=\textwidth]{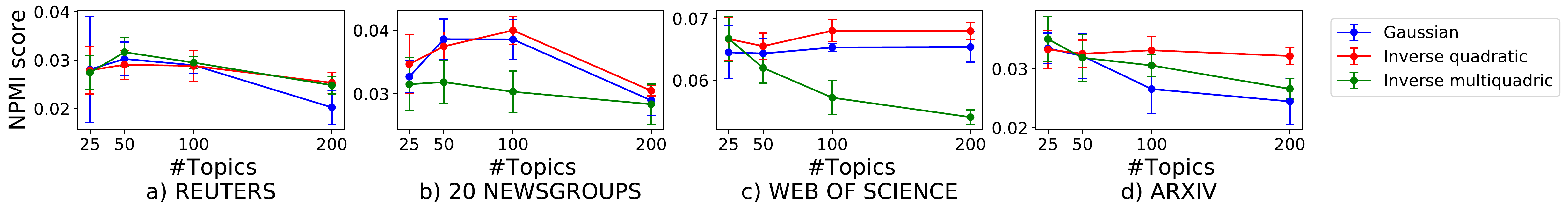}
	\caption{Topic coherence NPMI by \plsvvae\ with different RBFs (vary number of topics $Z$).}
	\label{fig:npmi_rbf}
\end{figure*}

\begin{figure*}
	\includegraphics[width=\textwidth]{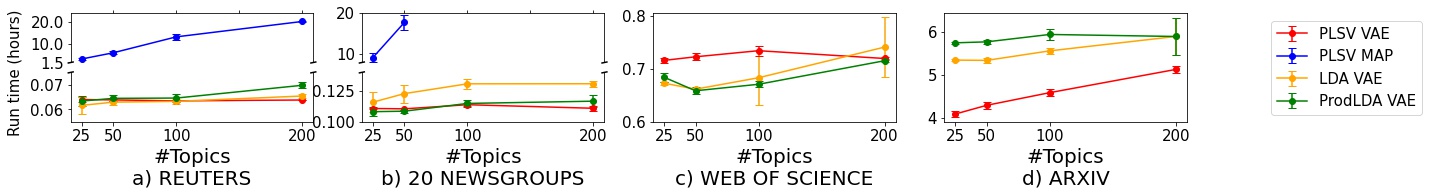}
	\caption{Running time comparison.}
	\label{fig:run_time}
\end{figure*}
}

Since our method is black box, we can quickly explore \plsvvae\ model with different assumptions. In this section, we show how different RBFs affect the performance of \plsvvae. Besides \plsvvae\ with Gaussian RBF, we implement another two variants of \plsvvae\ that uses two other RBFs: Inverse quadratic and Inverse multiquadric RBFs. We choose these two because, similar to Gaussian, they support the assumption that the $z$th topic proportion of document $n$ is high when document coordinate $x_n$ is close to topic coordinate $\phi_z$. For these model changes, we do not need to perform a mathematical rederivation, but we only need to change a few lines of code of \plsvvae\ (Gaussian). Figures \ref{fig:KNNRBF_KNN10} and \ref{fig:npmi_rbf} show the $k$-NN accuracy and topic coherence of \plsvvae\ with different RBFs. In general, \plsvvae\ with Gaussian or Inverse quadratic RBFs consistently produces good performance across datasets. In some cases, Inverse quadratic produces better results.
\subsection{Topic Examples and Running Time Comparison} \label{exp:topics}
To qualitatively evaluate the topics, in Figure \ref{fig:arxiv_Z50}, we show visualization and topic examples generated by \plsvvae\ (Inverse quadratic) on \arxiv. In the visualization, each black empty circle represents a topic that is associated with a list of top 10 words. We see that the topics are meaningful and reflect different research subdomains discussed in the \arxiv\ papers. For example, many topics are studied in the \textit{CS} domain such as ``graph, g, vertex, k'', ``model, data, use, method'', and ``logic, program, system''. For the \textit{Astro} domain, we have topics like ``galaxi, cluster, star'', and ''observ, ray, model, star''. Topics such as ``energi, nucleu, reaction'' and ``electron, energi, atom'' are discussed in the \textit{Nucl} domain. By allowing the semantics to be infused in the visualization space, users can now not only see the documents but also their topics. The joint nature of the model may lead to potential applications in different visual text mining tasks.


Finally, we show the running time of all the methods in Figure \ref{fig:run_time}. As expected, \plsvmap\ running on a single core is very slow and it fails to return any results on large datasets even after 24 hours of running. \plsvvae\ runs much faster. It only needs about 5 hours for 200 topics on the largest dataset \arxiv. For completeness, we also include the running time of \ldavae, and \plda. \plsvvae\ is as fast as these methods. In summary, \plsvvae\ can find good topics and visualization while it can scale well to large datasets, which will increase its usability in practice.


%% file: exp1.tex
\vspace{-0.1in}
We evaluate the effectiveness and efficiency of our proposed AEVB based inference method for visualization and topic modeling both quantitatively and qualitatively. We use four real-world public datasets from different domains including newswire articles, newsgroups posts and academic papers.

\noindent \textbf{Dataset Description}
\hide{
\begin{table}
	\caption{Dataset statistics. Descriptions given in Section \ref{exp:data}.}
	\label{tab:datasets}
	\centering
	\vspace{-0.15in}
	\scalebox{0.8}{
	\begin{tabular}{|l|r|r|r|} \hline
			& \#categories		& \#documents $N$ & \# words $V$\\ \hline
			\news					& 20	& 18251  &	 3248\\ \hline	
			\reuters		& 8	& 7674	 & 3000\\ \hline
			\wos				& 7 & 46985	 & 4000\\ \hline
			\arxiv 				& 7 & 598748	 & 5000\\ \hline
	\end{tabular}
	}
	\vspace{-0.2in}
\end{table}
}
\begin{compactitem}
	\item \reuters \footnote{{\url{http://ana.cachopo.org/datasets-for-single-label-text-categorization}}}: contains 7674 newswire articles from 8 categories \cite{2007:phd-Ana-Cardoso-Cachopo}.
    \item \news \footnote{{\url{https://scikit-learn.org/0.19/datasets/twenty_newsgroups.html}}}: contains 18251 newsgroups posts from 20 categories.
    \item \wos \footnote{{\url{https://data.mendeley.com/datasets/9rw3vkcfy4/6}}}: we use Web of Science WOS-46985 dataset \cite{kowsari2017HDLTex}. It contains the abstracts and keywords of 46,985 published papers from 7 research domains: CS, Psychology, Medical, ECE, Civil, MAE, and Biochemistry.
    \item \arxiv \footnote{{\url{http://zhang18f.myweb.cs.uwindsor.ca/datasets/}}}: contains the titles and abstracts of 598,748 research papers from arXiv. The papers are from 7 categories: Math, CS, Nucl, Stat, Astro, Quant, and Physics.
\end{compactitem}
We perform preprocessing by removing stopwords and stemming. The vocabulary sizes are 3000, 3248, 4000, and 5000 for \reuters, \news, \wos, and \arxiv\ respectively. Note that our problem is unsupervised and the ground-truth class labels are mainly used for evaluation.
Before detailing the experiment results, we describe the comparative methods.



\noindent \textbf{Comparative Methods.}
\hide{
\begin{table}
	\caption{Radial basis functions}
	\label{tab:rbfdistance}
	\centering
	\begin{tabular}{ | c || c| }
		\hline 
		RBF distance 			&  \\ \hline                        
		Gaussian 				&  $e^{-d^2} $ \\ \hline
		Inverse quadratic 		&  $\frac{1}{1+d^2} $\\\hline
		Inverse multiquadric 	& $\frac{1}{\sqrt{1+d^2}}$ \\ \hline
	\end{tabular}

\end{table}
}
We compare the following methods for inferring topics and visualization: 

\textbf{Joint approach:}
	\begin{compactitem}
		\item \plsvmap \footnote{{We use the implementation at \url{https://github.com/tuanlvm/SEMAFORE}}}: the original \plsv\ using MAP estimation with EM algorithm \cite{iwata2008probabilistic}.
		\item \plsvvae\ (Gaussian) [this paper]\footnote{{The implementation of our method can be found at \url{https://github.com/dangpnh2/plsv_vae}}}: we apply our proposed variational auto-encoder (VAE) inference to \plsv\ where Gaussian RBF is used. We write \plsvvae\ to refer to \plsvvae\ (Gaussian).
		\item \plsvvae\ (Inverse quadratic) and \plsvvae\ (Inverse multiquadric) [this paper]: these are \plsvvae\ models with Inverse quadratic and Inverse multiquadric RBFs. Since our method is black box, we can quickly implement these two models by just changing a few lines of code of \plsvvae\ (Gaussian) implementation. 

	\end{compactitem}
		
\textbf{Pipeline approach:} this is the approach of topic modeling followed by embedding of documents' topic proportions for visualization. We compare the above joint models with two pipeline models:
	\begin{compactitem}
		\item \mbox{\ldavae\ + \tsne}: topic modeling by \lda\footnote{{
				We use the implementation at \url{https://github.com/akashgit/autoencoding_vi_for_topic_models}\label{avitmnote}}} with VAE inference \cite{Srivastava2017AutoencodingVI}, then use \tsne \footnote{{We use the Multicore t-SNE implementation at \url{https://github.com/DmitryUlyanov/Multicore-TSNE}}} \cite{maaten2008visualizing} to visualize the documents' topic proportions.
		\item \mbox{\plda\ + \tsne}: similar to the above but we use \plda\footref{avitmnote} instead of \ldavae. 
	\end{compactitem}
In the next sections, we report the experiment results averaged across 10 independent runs. For \plsv\ models, we choose $\lambda=0.01, \gamma = 1, \varphi = \frac{\mN}{Z}$ that work well for large datasets in our experiments. We run \plsvmap\ with the number of EM iterations set to 200 and the maximum number of iterations for the quasi-Newton algorithm set to 10. Following AVITM, we set $H1 = H2 = 100$, the batch size to 256, the number of samples $L$ per document to 1, the learning rate to 0.002, and use dropout with probability $p = 0.6$. We use Adam as our optimizing algorithm. VAE based models are trained with 1000 epochs. All the experiments are conducted on a system with 64GB memory, an Intel(R) Xeon(R) CPU E5-2623 v3, 16 cores at 3.00GHz. The GPU in use on this system is NVIDIA Quadro P2000 GPU with 1024 CUDA cores and 5 GB GDDR5. 
\subsection{Classification in the Visualization Space}
\begin{figure*}
	\centering
	\includegraphics[width=\textwidth]{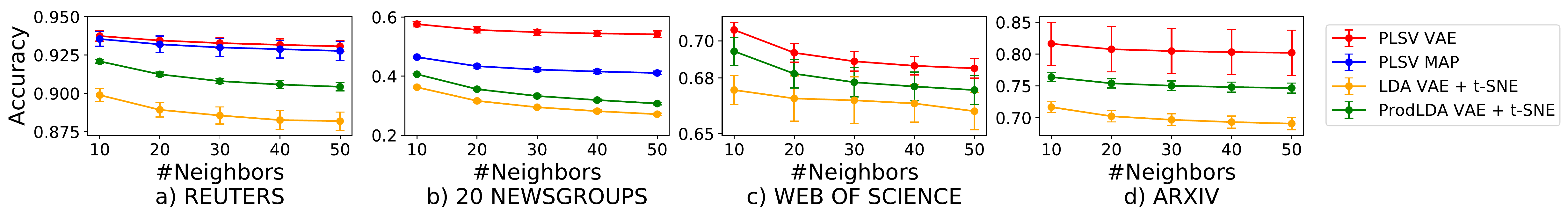}
	\vspace{-0.3in}
	\caption{$k$-NN accuracy in the visualization space with different number of nearest neighbors $k$ ($Z=50$ topics). For some settings, \plsvmap\ does not return any results even after 24 hours of running. 
	}
	\label{fig:KNN_Z50}
\vspace*{\floatsep}
	\includegraphics[width=\textwidth]{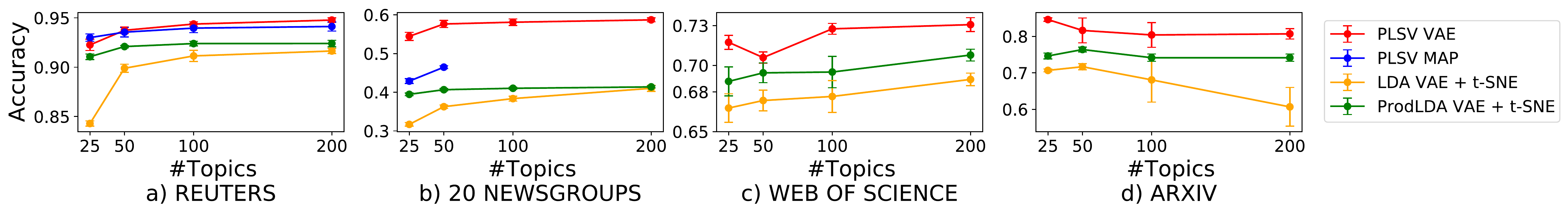}
	\vspace{-0.3in}
	\caption{$k$-NN accuracy in the visualization space with different number of topics $Z$ ($k=10$). For some settings, \plsvmap\ does not return any results even after 24 hours of running. 
	}
	\label{fig:KNN_KNN10}
\vspace*{\floatsep}
	\includegraphics[width=\textwidth]{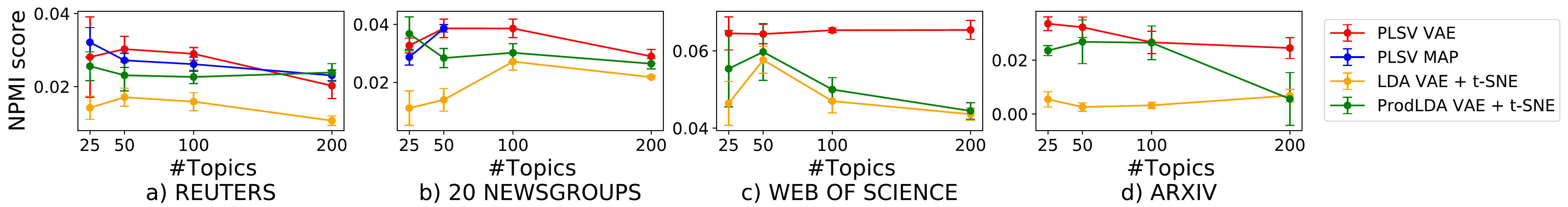}
	\vspace{-0.3in}
	\caption{Topic coherence based on NPMI with different number of topics $Z$. For some settings, \plsvmap\ does not return any results even after 24 hours of running.}
	\vspace{-0.15in}
	\label{fig:npmi}
\end{figure*}

We quantitatively evaluate the visualization quality by measuring the $k$-NN accuracy in the visualization space. This evaluation approach is also adopted in \tsne, \largevis, and the original \plsv. A $k$-NN classifier is used to classify documents using their visualization coordinates. A good visualization should group documents with the same label together and hence yield a high classification accuracy in the visualization space. Figures \ref{fig:KNN_Z50} and \ref{fig:KNN_KNN10} show $k$-NN accuracy of all methods on each dataset, for varying number of nearest neighbors $k$ and number of topics $Z$. For some settings, we do not show \plsvmap's performance as it does not return any results even after 24 hours of running. We can see that \plsvvae\ consistently achieves the best result, except for 25 topics on \reuters\ (Figure \ref{fig:KNN_KNN10}a) where it produces a comparable result with \plsvmap. These results show that the joint approach outperforms the pipeline approach and VAE inference may help improve the visualization quality of \plsv. To verify this qualitatively, in Section \ref{exp:visualization}, we show some visualization examples of all methods across datasets. Note that in this section, we show the accuracy of \plsvvae\ with Gaussian RBF. In Section \ref{exp:rbf}, we present the performance of \plsvvae\ with different RBFs.


%% file: conclusion.tex
\section{Conclusion}
We propose, to the best of our knowledge, the first fast AEVB based inference method for jointly learning topics and visualization. In our approach, we design a decoder that includes a normalized RBF network connected to a linear neural network. These networks are parameterized by topic coordinates and word probabilities, ensuring that they are shared across all documents. Due to our method's black box nature, we can quickly experiment with different RBFs with minimal reimplementation effort. Our extensive experiments on four real-world large datasets show that \plsvvae\ runs much faster than \plsvmap\ while gaining better visualization quality and comparable topic coherence.